\documentclass[11pt,a4paper]{article}
\usepackage[hyperref]{emnlp-ijcnlp-2019}
\usepackage{times}
\usepackage{latexsym}

\usepackage{url}

\usepackage{soul}
\usepackage[utf8]{inputenc}
\usepackage{graphicx}
\usepackage{amsmath}
\usepackage{booktabs}
\usepackage{algorithm}
\usepackage{algorithmic}
\usepackage{indentfirst}

\usepackage{multirow}
\usepackage{array}

\usepackage{amsfonts}

\DeclareMathOperator*{\myhardamard}{\raisebox{-3.5pt}{\scalebox{1.44}{$\Lambda$}}}

\newcommand{\tabincell}[2]{\begin{tabular}{@{}#1@{}}#2\end{tabular}}

\aclfinalcopy %

\title{Low-Rank HOCA: Efficient High-Order Cross-Modal Attention for Video Captioning}

\author{Tao Jin, Siyu Huang\thanks{\;\,Corresponding author.}, Yingming Li, and Zhongfei Zhang \\
	College of Information Science \& Electronic Engineering \\
	Zhejiang University, China \\
	\{jint\_zju, siyuhuang, yingming, zhongfei\}@zju.edu.cn}

\date{}

\begin{document}
\maketitle
\begin{abstract}
  %This paper attacks the challenging task of video captioning which aims to generate descriptions for video data. In the recent literature, video captioning has benefited from the advantages of applying attention mechanism to the multi-modal features in video data. With attention mechanism, the features are allocated different weights according to the information of an individual modality. However, in this way, the interaction between different modalities is ignored, although it has an impact on calculating the attention weights. We argue that different modalities are able to provide complementary information to each other. To solve this problem, we propose a video captioning model with High-Order Cross-Modal Attention (HOCA) where the attention weights are computed based on the correlation tensor to capture the frame-level interaction of different modalities sufficiently. Furthermore, we introduce a low-rank version of HOCA which uses tensor decomposition to reduce the large space requirement of high-order correlation, leading to a practical and efficient implementation in real-world applications. Empirical studies on two standard datasets demonstrate the superiority of our proposed method.
  This paper addresses the challenging task of video captioning which aims to generate descriptions for video data. Recently, the attention-based encoder-decoder structures have been widely used in video captioning. In existing literature, the attention weights are often built from the information of an individual modality, while, the association relationships between multiple modalities are neglected. Motivated by this observation, we propose a video captioning model with High-Order Cross-Modal Attention (HOCA) where the attention weights are calculated based on the high-order correlation tensor to capture the frame-level cross-modal interaction of different modalities sufficiently. Furthermore, we novelly introduce Low-Rank HOCA which adopts tensor decomposition to reduce the extremely large space requirement of HOCA, leading to a practical and efficient implementation in real-world applications. Experimental results on two benchmark datasets, MSVD and MSR-VTT, show that Low-rank HOCA establishes a new state-of-the-art.

  %In existing literature, the attention weights are built from the information of an individual modality, while, the association relationships between multiple modalities are neglected.
  
  %While existing methods retrieve the attention weights of video features according to the information of an individual modality and neglect the association relationships between multiple modalities.

\end{abstract}

\section{Introduction}

Video captioning has drawn much attention from natural language processing and computer vision researchers \cite{venugopalan2014translating,bin2016bidirectional,ramanishka2016multimodal,zanfir2016spatio}. As videos typically consist of multiple modalities (image, motion, audio, etc.), video captioning is actually a multimodal learning task. The abundant information underlying in the modalities is much beneficial for video captioning models. However, how to effectively learn the association relationships between different modalities is still a challenging problem.

%With the fact that video data contain rich temporal information in addition to the semantic information, video understanding is arguably considered more challenging than image understanding.

\begin{figure}[!t]
	\centering
	\includegraphics[scale=0.8]{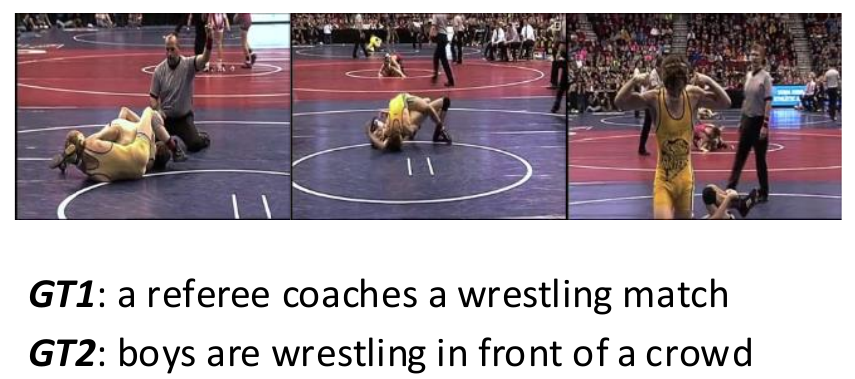}
	\caption{An example of video captioning, where "GT" denotes ground truth.}
	\label{fig:illustration}
\vspace{-0.4cm}
\end{figure}

%\vspace{-30pt}

In the context of deep learning based video captioning, multimodal attention mechanisms \cite{xu2017learning,hori2017attention,wang2018watch} are shown to help deliver superior performance. Bahdanau Attention \cite{bahdanau2014neural} is widely used to calculate the attention weights according to the features of individual modalities, since Bahdanau Attention is originally proposed for machine translation which can be considered as a unimodal task. However, video captioning is a multimodal task and different modalities are able to provide complementary cues to each other when calculating the attention weights. For example, as shown in Fig. \ref{fig:illustration}, a part of the video shows that two men are wrestling and another part shows that the referee coaches the match. With only the image modality, the model may pay equal attention to the frames with the competitors and referee. Considering the additional motion modality, the attention mechanism weighs more on one of them, resulting in a model with more focused attention.

Motivated by the above observations, in this paper we propose an attention mechanism called High-Order Cross-Modal Attention (HOCA) for video captioning, which makes a full use of the different modalities in video data by capturing their structure relationships. The key idea of HOCA is to consider the information of the other modalities when calculating the attention weights for one modality, different from Bahdanau Attention. In addition, we propose a low-rank version of HOCA which significantly reduces the computational complexity of HOCA. Specifically, the attention weights of HOCA are computed based on the similarity tensor between modalities, fully exploiting the correlation information of different modalities at each time step. Given the fact that the space requirement of the high-order tensor increases exponentially in the number of modalities and inspired by tensor decomposition, we adopt a low-rank correlation structure across modalities into HOCA to enable a good scalability to the increasing number of modalities. Such improvement largely reduces the algorithm complexity of HOCA with good results in empirical study. 

%we propose a cross-modal attention called high-order cross-modal attention (HOCA)

%The main contributions of this paper can be summarized as follow:

Our contributions can be summarized as:

(1) We propose High-Order Cross-Modal Attention (HOCA), which is a novel multimodal attention mechanism for video captioning. Compared with the Bahdanau Attention, HOCA captures the frame-level interaction between different modalities when computing the attention weights, leading to an effective multimodal modeling.

(2) Considering the scalability to the increasing number of modalities, we propose Low-Rank HOCA, which employs tensor decomposition to enable an efficient implementation of High-Order Cross-Modal Attention.

(3) Experimental results show that our method outperforms the state-of-the-art methods on video captioning, demonstrating the effectiveness of our method. In addition, the theoretical and experimental complexity analyses show that Low-Rank HOCA can implement multimodal correlation effeciently with acceptable computing cost.

\section{Related Work}

\subsection{Attention Mechanism}

The encoder-decoder structures have been widely used in sequence transformation tasks. Some models also connect the encoder and decoder through an attention mechanism\cite{bahdanau2014neural,luong2015effective}. In natural language processing (NLP), \cite{bahdanau2014neural} first proposes the soft attention mechanism to adaptively learn the context vector of the target keys/values. Such attention mechanisms are used in conjunction with a recurrent network. In the context of video captioning, there are many methods  \cite{hori2017attention,xu2017learning} using Bahdanau Attention to learn the context vector of the temporal features of video data, including the recent state-of-the-art method, hierarchically aligned cross-modal attentive network (HACA) \cite{wang2018watch}. Such method ignores the structure relationships between modalities when computing the attention weights. %However, our method captures the frame-level correlation between different modalities.

%To fully expoit the interaction information, HOCA captures the frame-level correlation between different modalities.

%Different from the existing work, HOCA further captures the frame-level correlation between different modalities.

%proposes a binary attention mechanism with modalities of image and motion, such mechanism utilizes the global feature of one modality to correlate the local features of the other modality. Different from the existing work mentioned above, HOCA further captures the frame-level correlation between different modalities when calculating the attention weights.      

%utilizes the global information of two modalities in video data to develop a binary attention mechanism. However, the approach does not capture the frame-level correlation, and, has a poor scalability to a large number of modalities.

%For example, SA \cite{venugopalan2014translating} first introduces the attention mechanisms into video captioning.

% or convolutional network \cite{gehring2017convolutional}. Recently, the multi-head attention and self attention \cite{vaswani2017attention} without any recurrent and convolutional networks demonstrate their effectiveness on many NLP tasks. 

\subsection{Video Captioning}

Compared with image data, video data have rich multimodal information, such as image, motion, audio, semantic object, text. Therefore, the key is how to utilize the information. In the literature of video captioning, \cite{xu2017learning,hori2017attention} propose the hierarchical multimodal attention, which selectively attends to a certain modality when generating descriptions. \cite{shen2017weakly,gan2017semantic} adopt multi-label learning with weak supervision to extract semantic features of video data. \cite{wang2018video} proposes optimizing the metrics directly with hierarchical reinforcement learning. \cite{chen2017video} extracts five types of features to develop the multimodal video captioning method and achieves promising results. \cite{wang2018reconstruction} performs the reconstruction of input video features by utilizing the output of the decoder, increasing the consistency of descriptions and video features. \cite{wang2018watch} proposes the hierarchically aligned multimodal attention (HACA) to selectively fuse both global and local temporal dynamics of different modalities. 

%However, they both ignore the interaction between the different modalities when calculating the attention weights.

%In terms of video question answering, \cite{yu2017end} uses a semantic word detector to tackle with the task, fully utilizing the semantic information with attention mechanism. \cite{xu2017video} proposes an encoder-decoder structure with a gradually refined attention mechanism. The existing work of video summarization only uses image features. \cite{zhang2016video} proposes an LSTM based structure for video summarization. \cite{mahasseni2017unsupervised} applies the adversarial learning to the task without supervision and achieves high performance. %\cite{zhou2018deep} adopts the reinforcement learning with self-defined metrics.  

%\cite{mahasseni2017unsupervised,zhou2018deep} introduce the unsupervised learning into the task, achieving superior results.

%\cite{mahasseni2017unsupervised} applies the adversarial learning to the task without supervision and achieves high performance. %\cite{zhou2018deep} adopts the reinforcement learning with self-defined metrics.  

%\cite{mahasseni2017unsupervised,zhou2018deep} introduce the unsupervised learning into the task, achieving superior results. 

None of the methods mentioned above utilizes the interaction information of different modalities to calculate the attention weights. Motivated by this observation, we present HOCA and Low-Rank HOCA for video captioning. Different from the widely used Bahdanau Attention, our methods fully exploits the video representation of different modalities and their frame-level interaction information when computing the attention weights.

We introduce our methods in the following sections, where Section \ref{background} introduces the details of Bahdanau Attention based multimodal video captioning (background), Section \ref{sec_att} gives the derivations of HOCA and Low-Rank HOCA, and the encoder-decoder structure which we propose for video captioning. Section \ref{sec_method} and \ref{sec_result} show the experimental settings and results of our methods.

\section{Multimodal Video Captioning}
\label{background}

Encoder-decoder structures combined with Bahdanau Attention are widely used in multimodal video captioning. Suppose that the number of input modalities is $n$, $I_l$ denotes the features of $l$-th modality, the space of $I_l$ is $\mathbb{R}^{d_l \times t_l}$ where $t_l$ denotes the temporal length and $d_l$ denotes the feature dimensions. The corresponding output is word sequence. 

\subsection{Encoder}
The input video is fed to multiple feature extractors, which can be pre-trained CNNs for classification tasks such as Inception-Resnet-v2\cite{szegedy2017inception}, I3D\cite{carreira2017quo}, VGGish\cite{hershey2017cnn}, each extractor corresponds to one modality. The extracted features are sent to Bi-LSTM \cite{hochreiter1997long} which has a capability to process sequence data, capturing the information from both forward and backward directions. The output of Bi-LSTM is kept.

\subsection{Decoder}
The decoder aims to generate word sequence by utilizing the features provided by the encoder and the context information. At time step $t$, the decoder output can be obtained with the input word $y_t$ and the previous output $h_{t-1}$,

\vspace{-5pt}
\begin{equation}
\label{1}
h_t = {\rm LSTM}(h_{t-1},y_t)
\end{equation}

\begin{figure*}[h]
	\centering
	\includegraphics[scale=0.5]{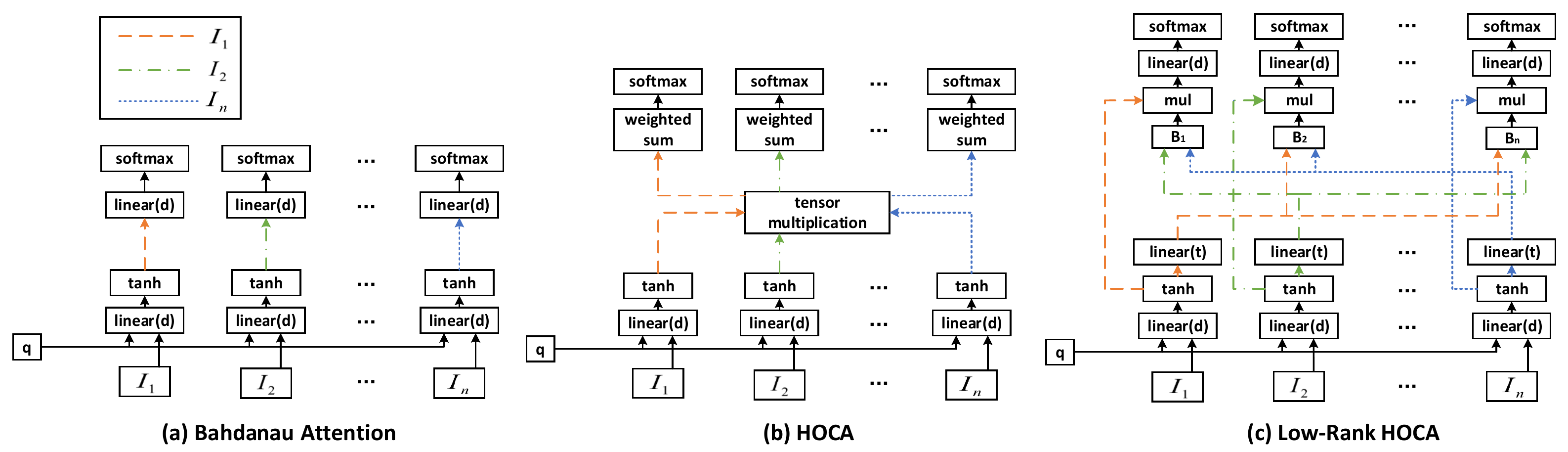}
	\vspace{-20pt}
	\caption{The attention mechanisms for multiple modalities where $n$ denotes the number of modalities. (a) shows Bahdanau Attention for different modalities; $I_l$ represents the $l$-th modality; linear(d) denotes the linear layer connected to the first dimension of $I_l$. (b) is the detailed structure of HOCA where the core module is ``tensor multiplication" defined by ourselves. (c) is Low-Rank HOCA, where linear(t) denotes the linear layer in temporal (second) dimension; ``mul" denotes multiplication operation; $B_l$ is defined in Eqn. \ref{16}.}
	\label{fig:hoca} 
\end{figure*}

We treat $h_t$ as query $q$, the features of different modalities are allocated weights with Bahdanau Attention seperately as shown in Fig. \ref{fig:hoca}(a). The attention weights $\alpha_t^{I_l} \in \mathbb{R}^{t_l}$ of $l$-th modality can be obtained as follows:
\vspace{-5pt}

\begin{equation}
\label{2}
(\alpha_{t}^{I_l})_{r_l}=\frac{{\rm exp}\Big \{(e_{t}^{I_l})_{r_l} \Big \}}{\sum_{n=1}^{t_l} {\rm exp} \Big \{(e_{t}^{I_l})_n \Big \}}
\end{equation}

\vspace{-10pt}

\begin{equation}
\label{3}
(e_{t}^{I_l})_{r_l}=w_1^\mathbb{T} {\rm tanh}\Big [W_1 h_t + U_1 (I_l)_{r_l}+b_1 \Big ]
\end{equation}
\vspace{-15pt}

\noindent where $(I_l)_{r_l}$ denotes the $r_l$-th time step of $I_l$ and $(\alpha_t^{I_l})_{r_l}$ is the corresponding weight, we combine the attention weights and features, obtaining the context vector $\varphi_t(I_l)$ of $l$-th modality,

\vspace{-10pt}
\begin{equation}
\label{4}
\sum_{{r_l}=1}^{t_l} (\alpha_{t}^{I_l})_{r_l} =1
\end{equation}

\vspace{-10pt}

\begin{equation}
\label{5}
\varphi_t(I_l) = \sum_{{r_l}=1}^{t_l} (\alpha_{t}^{I_l})_{r_l} (I_l)_{r_l}
\end{equation}

\vspace{-5pt}

\noindent the context vectors of other modalities can be obtained in the same way. We then integrate the context vectors to predict the word as follows:

\vspace{-10pt}

\begin{equation}
\label{6}
p_t = {\rm softmax} \Big [W_p^h h_t + \sum_{i=1}^{n} W_p^{I_i} \varphi_t(I_i) +b_p \Big ]
\end{equation}

\vspace{-5pt}

\noindent where $n$ is the number of modalities, $p_t$ denotes the probability distribution of words at time step $t$.

\section{HOCA and Low-Rank HOCA}
\label{sec_att}

%In this section, we start by introducing the Bahdanau Attention which is widely used in the existing work of video captioning. 

Considering multimodal features, Bahdanau Attention and its variants process different modalities separately. In such situation, the interaction between different modalities is ignored. We propose HOCA and Low-Rank HOCA to excavate this information, in addition, the tensor decomposition used in Low-Rank HOCA reduces the complexity of the high-order correlation.

%\subsection{Bahdanau Attention}

%Bahdanau Attention is widely used in video understanding tasks. Given a query $q$, the video features have different attention weights at each time step. Suppose that we have a video modality $I$ with space $\mathbb{R}^{d \times t}$ of the features. We obtain the weights $\alpha$ with space $\mathbb{R}^{t}$ through the following equations:

%\begin{equation}
%\label{1}
%\alpha_{i}=\frac{exp\{e_{i} \}}{\sum_{n=1}^{t} exp\{e_{n} \}}
%\end{equation}

%\begin{equation}
%\label{2}
%e_{i}=w_1^\mathbb{T} tanh(W_1 q + U_1 I_i+b_1)
%\end{equation}

%\noindent where $\alpha_{i}$ represents the relative importance of the feature $I_i$ for the target result. It is related to the query $q$.

\subsection{High-Order Cross-Modal Attention (HOCA)}
\label{sec_hoca}

Different modalities in video data are able to provide complementary cues for each other and should be integrated effectively. Inspired by the correlation matrix which is widely used in the natural language understanding \cite{seo2016bidirectional}, we use a high-order correlation tensor to model the interaction between different modalities. Fig. \ref{fig:hoca}(b) shows the generalized form of the structure for $n$ modalities, the core module is ``tensor multiplication" defined by ourselves. After the nonlinear mapping similar to Eqn. \ref{3} with query $h_t$, the features of the $n$ modalities are in a $d$-dimensional common space. Note that, for convenience, we still use $I_l$ to represent the features of the $l$-th modality (instead of $I_{l,t}$) and omit $t$ which denotes the time step of the decoder in the following derivations of Section \ref{sec_hoca} and \ref{sec_lr}.

%Suppose that the temporal length of modality $I_l$ is $t_l$, thus the space of $I_l$ is $\mathbb{R}^{d \times t_l}$. 

Let $\alpha_{n}^{I_l}$ denote the target attention weights of modality ${I_l}$. We obtain $\alpha_{n}^{I_l}$ through the high-order correlation tensor $C_{n}$ between the $n$ modalities. $C_{n}$ can be obtained as below:  

\vspace{-20pt}

\begin{equation}
\label{7}
\begin{aligned}
(C_{n})_{r_1,...,r_n} &=  {\bf 1}_{d}  \Big [(I_1)_{r_1} \circ (I_2)_{r_2} \circ...\circ (I_n)_{r_n} \Big ] \\
&= {\bf 1}_{d} \Big [\myhardamard_{i=1}^n (I_i)_{r_i} \Big]
\end{aligned}
\end{equation}

\vspace{-30pt}

\begin{equation}
\label{8}
\begin{aligned}
C_{n} &= \bigotimes \Big \{I_1 , I_2 ,..., I_n \Big \} = \bigotimes_{i=1}^n I_i
\end{aligned} 
\end{equation}

\vspace{-5pt}

\noindent where the $(r_1,...,r_n)$-th entry of $C_{n}$ is the inner-product of the $r_1$-th column (time step) of $I_1$, $r_2$-th column (time step) of $I_2$,...,$r_n$-th column (time step) of $I_n$. $\circ$ denotes element-wise multiplication and $\Lambda$ denotes the element-wise multiplication $\circ$ for a sequence of tensors. $\bigotimes$ is the tensor product operator and we use it to define a new operation $\bigotimes\{,...,\}$ for the ``tensor multiplication" of multiple matrices. ${\bf 1}_{d}$ with space $\mathbb{R}^{d}$ consists of $1$. We use the vectors ${\bf 1}_{d}$ and ${\bf 1}_{t_i}$ to denote the summation operation along the $d$ and $t_i$ dimensions. In this situation, Eqn. \ref{7} and Eqn. \ref{8} are equivalent. The attention weights of $I_l$ can be calculated as below:

\vspace{-20pt}

%\begin{shrinkeq}{-1ex}
\begin{equation}
\label{9}
(\alpha_n^{I_l})_{r_l}  = \frac{{\rm exp} \Big \{\sum \Big [W_{n-1}^{I_l} \circ (C_{n}^{I_l})_{r_l} \Big ]\Big \}}{\sum_{o=1}^{t_l} {\rm exp}\Big \{\sum\Big [W_{n-1}^{I_l} \circ (C_{n}^{I_l})_o \Big ] \Big \}}
\end{equation}
%\end{shrinkeq}

\noindent where $(C_{n}^{I_l})_{r_l}$ is equal to $(C_{n})_{:,:,...,r_l,...,:,:}$, which is an ($n$-$1$)-order tensor and denotes the correlation values between the $r_l$-th time step of $I_l$ and different time steps of the other $n$-$1$ modalities. $W_{n-1}^{I_l}$ is also an ($n$-$1$)-order tensor, which has the same shape and denotes the relative importance of the correlation values. $\sum$ denotes the summation function for the high-order tensor. For simplicity, the ``weighted sum" module can be considered as a linear layer in multiple dimensions.

%\begin{table}[!h]
%	\renewcommand{\arraystretch}{1.2}
%	\caption{The needed storage between "tanh" and "softmax" of HOCA and low-rank HOCA in Fig. \ref{fig:hoca}. Suppose that the number of time steps is $80$ for each modality, the batch size is $50$, $d_{al}$ is 512, the low rank $h$ is set to $1$. The elements in the tensor are expressed by $float32$ type. M denotes MByte, G denotes GByte. The details of the calculation are shown in the supplementary materials\textsuperscript{\ref {web}}. Note that the total storage is calculated by multiplying the data listed in the table and the number of query steps.}
%	\label{table:size}
%	\centering
%	\begin{tabular}{c|c|c|c}
%		\hline
%		
%		n & 3 & 4 & 5 \\
%		\hline
%		HOCA   &722M  & 56G & 4516G  \\
%		
%		%low-rank HOCA  & 192M & 256M & 320M \\
%		low-rank HOCA &47M &63M & 79M \\
%		\hline
%		%Bahdanau ATT &24M & 32M &40M \\
%		%\hline
%	\end{tabular}
%\end{table}

\subsection{Low-Rank HOCA}
\label{sec_lr}

%As the number of the modalities $n$ increases, the space requirement of HOCA becomes more demanding as shown in Table \ref{table:size}. In addition, Table \ref{table:size} only exhibits the storage of one step; the total storage is calculated by multiplying the data listed in Table \ref{table:size} and the number of query steps. 

One of the main drawbacks of HOCA is the generation of high-order tensor, the size of the high-order tensor will increases exponentially with the number of modalities as $\prod_{i=1}^{n} t_i$, resulting in a lot of computation. Therefore, we implement the multimodal correlation between the different modalities in a more efficient way with low-rank approximation which has been widely used in the community of vision and language \cite{lei2015molding,liu2018efficient,yu2017multi}. Following the Eqn. \ref{7} and \ref{8}, we rewrite $(C_{n}^{I_l})_{r_l}$ as\textsuperscript{\ref {web}}:

\vspace{-10pt}

\begin{equation}
\label{10}
\begin{aligned}
(C_{n}^{I_l})_{r_l} &= \bigotimes \Big \{(I_l)_{r_l} \circ I_1 , ... , I_{l-1} , I_{l+1} ,..., I_n \Big \} \\
&= \bigotimes_{i=1,\not=l}^n (I_l)_{r_l} \star I_i
\end{aligned}
\end{equation}

Element-wise multiplication operator $\circ$ is used for vector $(I_l)_{r_l}$ and matrix $I_1$. Each column of matrix $I_1$ multiplies vector $(I_l)_{r_l}$. $\star$ denotes that $(I_l)_{r_l}$ is multiplied ($\circ$) only when $i=1$. We assume a low-rank factorization of the tensor $W_{n-1}^{I_l}$. Specifically, $W_{n-1}^{I_l}$ is decomposed into a sum of $k$ rank $1$ tensors,

\vspace{-10pt}

\begin{equation}
\label{11}
W_{n-1}^{I_l}=\sum_{j=1}^{k}  \bigotimes_{i=1,\not=l}^n w_j^{I_i}
\end{equation}

\vspace{-10pt}

\noindent where the space of $w_j^{I_i}$ is $\mathbb{R}^{1 \times t_i}$. Note that we set $k$ to a constant value and use the recovered low-rank tensor to approximate $W_{n-1}^{I_l}$. The numerator in Eqn. \ref{9} can be further derived as: 

\vspace{-10pt}

\begin{equation}
\label{12}
\begin{aligned}
\sum \Big [W_{n-1}^{I_l} \! \circ \! (C_{n}^{I_l})_{r_l} \Big ] \! = \! \sum \!\Big [\sum_{j=1}^{k} \bigotimes_{i=1,\not=l}^n w_j^{I_i} \\ \circ  
\bigotimes_{i=1,\not=l}^n (I_l)_{r_l} \star I_i \Big ]
\end{aligned}
\end{equation}

\noindent Since $I_i$ is a matrix and $w_j^{I_i}$ is a vector, we directly multiply $I_i$ and corresponding $w_j^{I_i}$ \footnote{The details of the propositions are shown in the supplementary materials.\label{web}}. Each row of the matrix $I_i$ multiplies vector $w_j^{I_i}$, 

%\begin{shrinkeq}{-1ex}
%	\begin{equation}
%	\label{8}
%	\begin{aligned}
%	{\rm Eqn}. \ref{7} = \sum_{i=1}^{h} (\sum \otimes \{(I_l)_{r_l} \odot (I_1 \odot w_i^{I_1})  , 
%	..., (I_n \odot w_i^{I_n} )\})
%	\end{aligned}
%	\end{equation}
%\end{shrinkeq}
\vspace{-10pt}

\begin{equation}
\label{13}
\begin{aligned}
\sum \!\Big [W_{n-1}^{I_l} \! \circ \! (C_{n}^{I_l})_{r_l} \Big ] \! = \! \sum_{j=1}^{k} \! \Big [\sum \! \bigotimes_{i=1,\not=l}^n \! (I_l)_{r_l} \star I_i \circ w_j^{I_i} \Big ]
\end{aligned}
\end{equation}

\begin{figure*}[!h]
	\centering
	\includegraphics[scale=0.53]{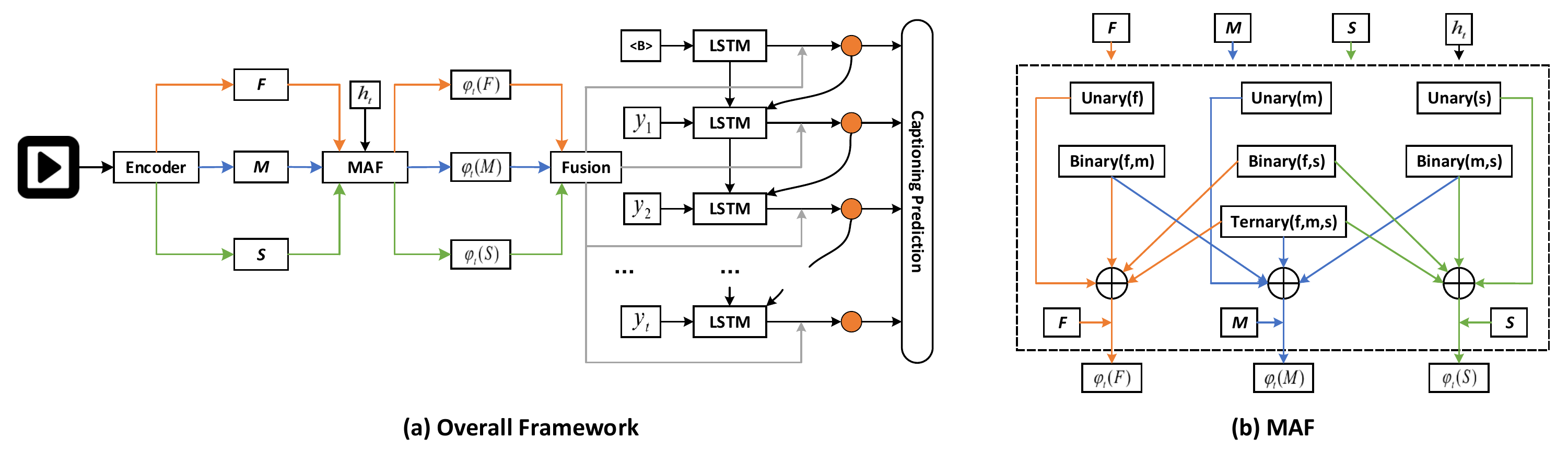}
	\vspace{-10pt}
	\caption{(a) is the overall framework of video captioning where F, M, S represent image, motion, and audio, respectively; $h_t$ denotes the query at time step $t$. We predict words at all the time steps of the decoder with multiple attentive fusion (MAF) module. (b) is the detailed structure of MAF module, each modality has three types of attention weights (unary, binary, and ternary),  where ``Ternary" denotes that the number of modalities is $3$ in HOCA and Low-Rank HOCA, ``Binary" and ``Unary" correspond to $2$ and $1$. The unary attention is equal to Bahdanau Attention. In addition, ``Binary(f,m)" denotes the binary attention weights with image and motion. We utilize trainable parameters to determine the importance of these attention weights when integrating them.}
	\label{fig:overall_famework} 
\end{figure*}

\noindent where we first apply the tensor multiplication $\bigotimes$ to correlating the different time steps of all modalities. During the process, we sum in the $d$ dimension to obtain the elements of the high-order tensor. Second, we sum (inner $\sum$) all the elements. For convenience, we change the operation order, we first sum in the temporal dimension, then sum in the $d$ dimension\textsuperscript{\ref {web}}. Letting $(I_i^{'})_{j}$ denote the global information of $I_i$ with importance factor $w_j^{I_i}$, 

\vspace{-10pt}

\begin{equation}
\label{14}
(I_i^{'})_{j} =  (I_i \circ  w_j^{I_i}) {\bf 1}_{t_i} = I_i (w_j^{I_i})^{\mathbb{T}}
\end{equation}

\noindent Eqn. \ref{12} can be further derived as below:

%and then summing (inner $\sum$) them.

\vspace{-10pt}

\begin{equation}
\label{15}
\begin{aligned}
\sum \Big [ W_{n-1}^{I_l} \!\circ \!(C_{n}^{I_l})_{r_l} \Big ] \!=\!{\bf 1}_{d} \Big [(I_l)_{r_l} \!\circ \! B_l \Big ]
\end{aligned}
\end{equation}

\vspace{-20pt}

\begin{equation}
\label{16}
B_l =\sum_{j=1}^{k} \myhardamard_{i=1,\not=l}^n  (I_i^{'})_{j}
\end{equation}

Due to the different information carried by the elements of the feature, we use a linear layer $w_{I_l}$ to replace ${\bf 1}_{d}$,

\vspace{-10pt}

\begin{equation}
\label{17}
(\alpha_n^{I_l})_{r_l}  = \frac{{\rm exp} \Big \{ w_{I_l} \Big [(I_l)_{r_l} \circ B_l \Big ] \Big \}}{\sum_{o=1}^{t_l} {\rm exp} \Big \{ w_{I_l} \Big [(I_l)_{o} \circ B_l \Big ] \Big \}}
\end{equation}

\noindent The detailed structure of Low-Rank HOCA is shown in Fig. \ref{fig:hoca}(c).

\subsection{Complexity Analysis}
\label{sec_analysis}

We analyze the space complexity of Bahdanau Attention, HOCA, and Low-Rank HOCA in Fig. \ref{fig:hoca}, focusing on the trainable variables and the output of each layer. For convenience, we start by calculating from the output of $tanh$ layer, since the front structures of three methods are same.

\vspace{0.2cm}

{\bf Bahdanau Attention} The size of the trainable variable in second $linear(d)$ layer is $d \times 1$, and the size of the output is $\sum_{i=1}^{n} t_i$. The space complexity is $O(\sum_{i=1}^{n} t_i +d)$. 

%and the complexity is $O(\sum_{i=1}^{n} t_i)$.
\vspace{0.2cm}

{\bf HOCA} The size of the output of tensor multiplication is $\prod_{i=1}^{n} t_i$. The size of the trainable variables and corresponding output in $weighted$ $sum$ layer is $\sum_{i=1}^{n} (\frac{\prod_{j=1}^{n} t_j}{t_i}) + \sum_{i=1}^{n} t_i$. The space complexity is $O(\prod_{i=1}^{n} t_i + \sum_{i=1}^{n} (\frac{\prod_{j=1}^{n} t_j}{t_i}) + \sum_{i=1}^{n} t_i)$. 

%and the complexity is $O(\prod_{i=1}^{n} t_i)$.

\vspace{0.2cm}
{\bf Low-Rank HOCA} The rank is set to $k$. The size of the trainable variable and corresponding output in $linear(t)$ layer is $\sum_{i=1}^{n} t_i \times k + n \times k \times d$. The size of the output in $B_l$ and $mul$ layer is $n \times d + \sum_{i=1}^{n} t_i \times d$. The size of the trainable variable and the corresponding output in second $linear(d)$ layer is $d + \sum_{i=1}^{n} t_i$. The space complexity is $O(\sum_{i=1}^{n} t_i \times (k+d+1) + (n\times k +n +1) \times d)$. 

\vspace{0.2cm}

%and the complexity is $O(\sum_{i=1}^{n} t_i)$.

Bahdanau Attention and Low-Rank HOCA both scale linearly in the number of modalities while HOCA scales exponentially. Therefore, HOCA will have explosive complexity when the number of modalities is big, and Low-Rank HOCA can solve this problem effectively.

\begin{table*}[!t]
	\renewcommand{\arraystretch}{1.2}
	\setlength\tabcolsep{3.0pt}
	\caption{Evaluation results of our proposed models}
	\label{table:compa}
	\centering
	\begin{tabular}{c|c|c|c|c|c|c|c|c}
		\hline
		%\bfseries First & \bfseries Next\\
		\multirow{2}*{Method} & \multicolumn{4}{c|}{MSVD} & \multicolumn{4}{c}{MSR-VTT} \\
		\cline{2-9}
		& BLEU4
		& ROUGE
		& METEOR
		& CIDEr
		& BLEU4
		& ROUGE
		& METEOR
		& CIDEr\\
		
		\hline\hline
		%only image(F) &50.6 &70.6 &34.3 &83.7 &41.9 &61.2 &29.1 &47.2 \\
		%only motion(M) &50.6 &70.6 &34.3 &83.7 &41.9 &61.2 &29.1 &47.2 \\
		
		%\hline
		%only audio &50.6 &70.6 &34.3 &83.7 &41.9 &61.2 &29.1 &47.2 \\
		%HOCA-UB(FM) &50.6 &70.6 &34.3 &83.7 &41.9 &61.2 &29.1 &47.2 \\
		%HOCA-UB(FS) &50.6 &70.6 &34.3 &83.7 &41.9 &61.2 &29.1 &47.2 \\
		%HOCA-UB(MS) &50.6 &70.6 &34.3 &83.7 &41.9 &61.2 &29.1 &47.2 \\
		HOCA-U &50.6 &70.6 &34.3 &83.7 &41.9 &61.2 &29.1 &47.2 \\
		HOCA-B &49.6 &70.8 &34.3 &84.3 &42.2 &61.1 &29.1 &47.6 \\
		L-HOCA-B &50.1 &70.4 &34.8 &84.7 &42.4 &61.4 &29.0 &47.9 \\
		HOCA-T &50.9 &71.2 &34.8 &84.8 &42.4 &61.5 &29.0 &48.1 \\
		L-HOCA-T &51.3 &71.5 &34.7 &84.8 &43.2 &61.5 &29.1 &48.4 \\
		HOCA-UB &50.6 &71.1 &34.0 &85.7 &43.1 &61.5 &29.0 &48.0 \\
		L-HOCA-UB &51.7 &71.9 & 34.7 &\bf 86.1 &43.5 &62.3 &29.2 &48.5  \\
		HOCA-UBT &52.3 &71.9 &\bf 35.5 & 85.3&43.9 &62.0 &29.0 &49.3  \\
		L-HOCA-UBT&\bf 52.9 &\bf 72.0 &\bf 35.5 &\bf 86.1 &\bf 44.6 &\bf 62.6 &\bf 29.5 &\bf 49.8 \\
		
		\hline
	\end{tabular}
\vspace{-0.2cm}
\end{table*}

\subsection{Video Captioning with HOCA and Low-Rank HOCA} 
In this section, we mainly introduce our encoder-decoder structure combined with high-order attention (HOCA and Low-Rank HOCA). As shown in Fig. \ref{fig:overall_famework}(a), the features of different modalities, i.e. image(F), motion(M), audio(S) are extracted in the encoder. These features are sent to the decoder for generating words. The ``MAF" module performs HOCA and Low-Rank HOCA for the features of different modalities with $h_t$. As shown in Fig. \ref{fig:overall_famework}(b), each modality has three types of attention weights, i.e. unary, binary, ternary, which denote the different number (i.e. $1$,$2$,$3$)  of modalities applied to the HOCA and Low-Rank HOCA (note that unary attention is equal to Bahdanau Attention). In some cases, not all the modalities are effective, for example, the binary attention weights of image and motion are more accurate for the salient videos.

We use $\alpha_{t,3}^F$, $\alpha_{t,3}^M$, $\alpha_{t,3}^S$ to denote the ternary weights of three modalities at time step $t$, the calculation is shown as follows:

\vspace{-10pt}

\begin{equation}
\label{18}
\alpha_{t,3}^F,\alpha_{t,3}^M,\alpha_{t,3}^S = {\rm HOCA}(h_t, F, M, S)
\end{equation}

\noindent where $F$,$M$,$S$ denote the features of image, motion, audio, respectively, and HOCA can be replaced by Low-Rank HOCA. We obtain the binary and unary weights in the same way. For different attention weights, we utilize trainable variables to determine their importance. We take the image modality as an example, the fusion weights are obtained as follows:

\vspace{-10pt}

\begin{equation}
\label{19}
\alpha_t^F \! = \! \rm softmax(\theta_1 \alpha_{t,1}^F \!+ \!\theta_2 \alpha_{t,2}^{F,M} \! + \! \theta_3 \alpha_{t,2}^{F,S} \! + \! \theta_4 \alpha_{t,3}^F) \\
\end{equation}

\noindent where $\theta_{1-4}$ are trainable variables, $\alpha_{t,1}^F$ denotes the unary weights, $\alpha_{t,2}^{F,M}$ and $\alpha_{t,2}^{F,S}$ denote the binary weights with motion and audio, respectively. As we obtain the attention weights of three modalities, we can calculate the context vectors $\varphi_t(F)$, $\varphi_t(M)$, $\varphi_t(S)$ following the Eqn. \ref{4} and \ref{5}.

To further determine the relative importance of multiple modalities, we perform a hierarchical attention mechanism for the context vectors in the ``Fusion" module. The attention weights of image modality are calculated as follows:

\vspace{-10pt}

\begin{equation}
\label{20}
\beta_t^F = \frac{{\rm exp}(e_t^F)}{\sum_{k=\{F,M,S\}} {{\rm exp}(e_t^k)}}
\end{equation} 

\vspace{-10pt}

\begin{equation}
\label{21}
e_t^F = w_e^\mathbb{T} {\rm tanh} \Big [W_e h_t + U_e \varphi_t (F) + b_e \Big ]
\end{equation}

\noindent $\beta_t^M$ and $\beta_t^S$ are obtained in the same way. We integrate the context vectors and corresponding weights to predict word as follows:

\begin{equation}
\label{22}
p_t = {\rm softmax} \Big [W_p^h h_t + \sum_k \beta_t^k D_t (k) + b_p \Big ]
\end{equation}

\vspace{-20pt}

\begin{equation}
\label{23}
D_t(k) = W_p^k \varphi_t (k), k=\{F,M,S\}
\end{equation}

The optimization goal is to minimize the cross-entropy loss function defined as the accumulative loss from all the time steps:

\vspace{-10pt}

\begin{equation}
\label{24}
L = -\sum_{i=1}^{T} {\rm log} (p_t^*|p_1^*,...,p_{t-1}^*,V)
\end{equation}

\noindent where $p_t^*$ denotes the probability of the ground-truth word at time step $t$, $T$ denotes the length of description, $V$ denotes the original video data.

%HOCA-UB &52.6 & 71.0 &34.7 &86.1 &44.4 &62.1 &29.0 &48.6 \\

%The processed features are sent to "CM-ATT" (cross-modal attention) module, which is shown in Fig. \ref{fig:overall_famework}(c). The output are the attention context vectors of the three modalities, followed by the "Fusion" module similar to that in \cite{hori2017attention} to obtain the final representation. We predict words at each time step,

%\begin{shrinkeq}{-1.2ex}
%	\begin{equation}
%	\label{13}
%	p_t={\rm softmax}(W_p F_{f}(h_t,\varphi_t(f),\varphi_t(m), \varphi_t(s) ))
%	\end{equation}
%\end{shrinkeq}

%\noindent where $F_{f}$ represents the "Fusion" module. 

\section{Experimental Methodology}
\label{sec_method}

\subsection{Datasets and Metrics}
We evaluate video captioning on two standard datasets, MSVD \cite{chen2011collecting} and MSR-VTT \cite{xu2016msr}, which are both provided by Microsoft Research, with several state-of-the-art methods. MSVD includes 1970 video clips. The time length of a video clip is about 10 to 25 seconds and each video clip is annotated with about 40 English sentences. MSR-VTT has 10000 video clips; each clip is annotated with 20 English sentences. We follow the commonly used protocol in the previous work and use four common metrics in the evaluation, including BLEU4, ROUGE, METEOR, and CIDEr.

\subsection{Preprocessing and Experimental Setting}

%\cite{szegedy2017inception} \cite{carreira2017quo} \cite{hershey2017cnn}

We sample video data to $80$ frames for extracting image features. For extracting motion features, we divide the raw video data into video chunks centered on $80$ sampled frames at the first step. Each video chunk includes $64$ frames. For extracting audio features, we obtain the audio file from the raw video data with FFmpeg. For both datasets, we use a pre-trained Inception-ResNet-v2 \cite{szegedy2017inception} to extract image features from the sampled frames and we keep the activations from the penultimate layer. In addition, we use a pre-trained I3D \cite{carreira2017quo} to extract motion features from video chunks. We employ the activations from the last convolutional layer and implement a mean-pooling in the temporal dimension. We use the pre-trained VGGish \cite{hershey2017cnn} to extract audio features. On MSRVTT, we also utilize the glove embedding of the auxiliary video category labels to initialize the decoder state.

The hidden size is $512$ for all LSTMs. The attention layer size for image, motion, audio attention is also $512$. The dropout rate for the input and output of LSTM decoder is $0.5$. The rank is set to $1$. In the training stage, we use Adam \cite{kingma2014adam} algorithm to optimize the loss function; the learning rate is set to $0.0001$. In the testing stage, we use beam-search method with beam-width $5$. We use a pre-trained word2vec embedding to initialize the word vectors. Each word is represented as a $300$-dimension vector. Those words which are not in the word2vec matrix are initialized randomly. All the experiments are done on $4$ GTX 1080Ti GPUs.

\begin{figure*}[t]
	%\vspace{-5mm}
	\centering
	\includegraphics[scale=0.75]{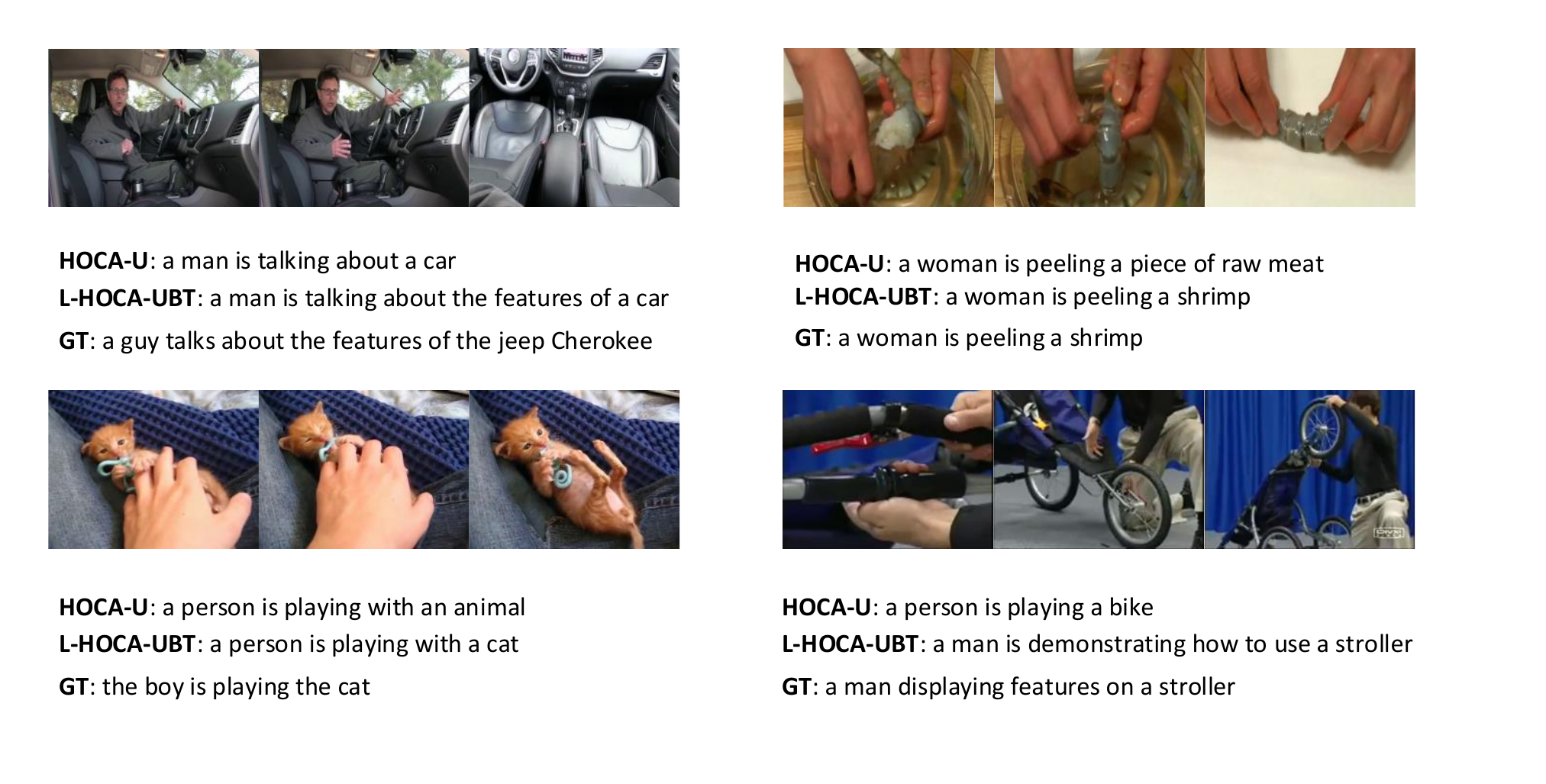}
	\caption{Qualitative results of Low-Rank HOCA}
	\label{fig:qualitative_result}
\vspace{-0.3cm}
\end{figure*}

\section{Experimental Results}
\label{sec_result}

\subsection{Impact of Cross-Modal Attention}

Table \ref{table:caption_result} shows the results of different variants of HOCA and Low-Rank HOCA. HOCA-U, HOCA-B, and HOCA-T denote the model with only unary, binary, and ternary attention, respectively. HOCA-UB and HOCA-UBT denote the models with original HOCA and more types of attention mechanisms. The prefix ``L-HOCA" denotes the model with Low-Rank HOCA.

It is observed that the model with only one type of  attention mechanism(U, B, or T) in the decoder achieves relatively bad results on both datasets. However, when we combine them, the performances are significantly improved on metrics, especially ROUGE and CIDEr. HOCA-UBT and L-HOCA-UBT with a combination of unary, binary, and ternary attention achieve relatively promising results on all the metrics. We argue that HOCA-UBT and L-HOCA-UBT can learn appropriate ratios of all types of attention mechanisms based on the specific video-description pairs, while other variants only focus on one or two types. In addition, the models of low-rank version (L-HOCA-UB and L-HOCA-UBT) have better metrics than the models of original version (HOCA-UB and HOCA-UBT). On the one hand, we utilize $w_{I_l}$ to replace ${\bf 1}_d$ in Eqn. \ref{17}, fully mining the different information carried by the elements of the feature, on the other hand, the low-rank approximation is effective.%\footnote{The learning curves are shown in the supplementary materials.\label{curve}}.  

%\begin{table}[!h]
%	\renewcommand{\arraystretch}{1.1}
%	\setlength\tabcolsep{6.0pt}
%	\caption{Evaluation results of different binary association on MSRVTT}
%	\label{table:binary}
%	\centering
%	\begin{tabular}{c|c|c|c|c}
%		\hline
%		Method & B & R & M & C\\
%		\hline
		
%		HOCA-U & 41.9 & 61.2 & 29.1 & 47.2\\
%		L-HOCA-UB(F+M) &42.1 &61.4 &29.0 &47.8 \\
%		L-HOCA-UB(F+S) &42.0 &61.3 &28.8 &47.3 \\
%		L-HOCA-UB(M+S) &41.7 &61.3 &29.0 &47.5 \\
%		L-HOCA-B & 42.4 & 61.4 &29.0 & 47.9\\
%		\hline
%	\end{tabular}
%	\vspace{-0.3cm}
%\end{table}

%HOCA-B reported above combines three binary associations. We add some experiments on MSRVTT to evaluate the performances of different binary associations. The results are shown in Table \ref{table:binary}, where L-HOCA-UB(F+M) denotes that image and motion modalities use binary attention, while audio modality use unary attention. L-HOCA-UB(F+S) and L-HOCA-UB(M+S) are similar. Generally, the metrics of single binary association are between those of HOCA-U and L-HOCA-B, but the differentials are quite small. In addition, among three binary associations, L-HOCA-UB(F+M) achieves relatively good results.

\begin{table}[!h]
	\renewcommand{\arraystretch}{1.2}
	\setlength\tabcolsep{6pt}
	\caption{Evaluation results of video captioning, where B, R, M, C denote BLEU4, ROUGE, METEOR, CIDEr, respectively, and Ours denotes L-HOCA-UBT.}
	\label{table:caption_result}
	\centering
	\begin{tabular}{p{1cm}|p{1.9cm}<{\centering}|p{0.55cm}<{\centering}|p{0.55cm}<{\centering}|p{0.55cm}<{\centering}|p{0.55cm}<{\centering}}
		\hline
		Dataset&Method & B & R & M & C\\
		\hline
		\multirow{7}{*}{\tabincell{c}{MSR-\\VTT}}
		&RecNet &39.1 & 59.3 &26.6 &42.7 \\
		&HRL &41.3 & 61.7 &28.7 &48.0\\
		%\cline{2-6}
		%&M-to-M &40.8 &60.2 &28.8 & 47.1 \\
		&Dense Cap & 41.4 &61.1 & 28.3 &48.9 \\
		&HACA & 43.5 & 61.8 & \bf 29.5 & 49.7 \\
		&MM-TGM & 44.3 & - &  29.3 & 49.2\\
		
		%\cline{2-6}
	
		\cline{2-6}
		
		&Ours &\bf 44.6 &\bf 62.6 &\bf 29.5 &\bf 49.8 \\

		\hline
		\hline
		\multirow{7}{*}{\tabincell{c}{MSVD}}
		&SCN  &50.2 &- & 33.4& 77.7\\
		&TDDF &45.8 & 69.7 &33.3&73.0\\

		&LSTM-TSA & 52.8&-&33.5&74\\
		&RecNet &52.3&69.8&34.1&80.3\\
		&MM-TGM &48.7 & - & 34.3 & 80.4\\
		\cline{2-6}
		
		&Ours &\bf 52.9 &\bf 72.0 &\bf 35.5 &\bf 86.1\\
		\hline
	\end{tabular}
\vspace{-0.4cm}
\end{table}

\subsection{Compared with the State-of-the-art}
Table \ref{table:caption_result} shows the results of different methods on MSR-VTT and MSVD, including ours (L-HOCA-UBT), and some state-of-the-art methods, such as LSTM-TSA \cite{pan2017video}, TDDF \cite{zhang2017task}, SCN \cite{gan2017semantic}, MM-TGM \cite{chen2017video}, Dense Caption \cite{shen2017weakly}, RecNet \cite{wang2018reconstruction}, HRL \cite{wang2018video},HACA\cite{wang2018watch}. 

From Table \ref{table:caption_result}, we find that Ours(L-HOCA-UBT) shows competitive performances compared with the state-of-the-art methods. On MSVD, L-HOCA-UBT has outperformed SCN, TDDF, RecNet, LSTM-TSA, MM-TGM, on all the metrics. In particular, L-HOCA-UBT achieves $86.1\%$ on CIDEr, making an improvement of $5.7\%$ over MM-TGM. On MSR-VTT, we have the similar observation, L-HOCA-UBT has outperformed RecNet, HRL, MM-TGM, Dense Caption, and HACA on all the metrics.

\begin{table}[!h]
	\renewcommand{\arraystretch}{1.1}
	\setlength\tabcolsep{2.0pt}
	\caption{Computing cost of different methods, where the ``space" denotes the memory space requirement and the ``training time" denotes the total time for training. We evaluate them on MSR-VTT. Note that the metrics belong to the whole model, not only the attention module.}
	\label{table:complexity}
	\centering
	\begin{tabular}{c|c|c}
		\hline
		Method & Space (G) & Training Time (s)\\
		\hline

		HOCA-U & 4.1 & 16418  \\
		HOCA-UBT &9.7 & 37040 \\
		L-HOCA-UBT &5.6 & 24615\\
		\hline
	\end{tabular}
\vspace{-0.3cm}
\end{table}

%\begin{figure}[!t]
%	\centering
%	\includegraphics[scale=0.7]{figs/quality2.pdf}
%	\caption{Visualization of the attention weights in "Fusion" module on video captioning where red bar denotes image, green and blue bars correspond to motion and audio.}
%	\label{fig:quality2}
%\end{figure}

%The results of video captioning and question answering are shown in the first row. The results of video summarization are shown in the second and the third row.

\vspace{-10pt}

\subsection{Computing Cost}
\label{sec_com}
The theoretical complexity of different attention mechanisms is illustrated in Section \ref{sec_analysis}. In practice, we utilize the experimental settings mentioned above, the batch size and the maximum number of epochs are set to $25$ and $100$, respectively. the training time and memory space requirement are shown in Table \ref{table:complexity}. We can find that L-HOCA-UBT has smaller space requirement and less time cost than HOCA-UBT, in addition, the computing cost of L-HOCA-UBT is close to that of HOCA-U (Bahdanau Attention). The results demonstrate the advantage of Low-rank HOCA.

%We also evaluate the training efficiency and show the results of HOCA-U, HOCA-UBT, L-HOCA-UBT, in Table \ref{table:complexity}, the time cost of Low-rank HOCA is less than HOCA, which demonstrates the advantage of Low-rank HOCA again.  

%which demonstrates the advantage of Low-rank HOCA again.

\subsection{Rank Setting}
\begin{figure}[h]
	\centering
	\includegraphics[scale=0.5]{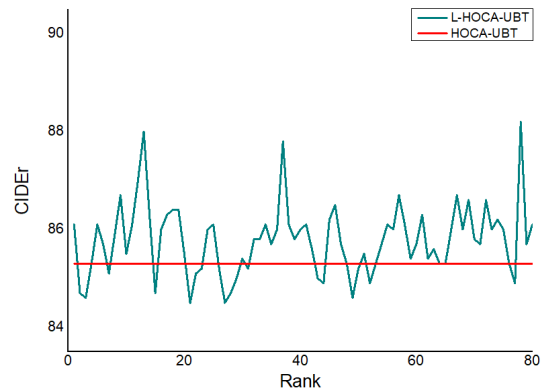}
	\caption{The performance of different rank values}
	\label{fig:rank}
%\vspace{0.3cm}
\end{figure}

We also evaluate the impact of different rank values. We show the results on MSVD in Fig. \ref{fig:rank}. The red and green lines represent HOCA-UBT and L-HOCA-UBT, respectively. We find that the CIDEr of L-HOCA-UBT has slight fluctuations as the rank changes and a small value of rank can achieve competitive results with high efficiency.

%\vspace{-10pt}

\subsection{Qualitative Analysis}
Fig. \ref{fig:qualitative_result} shows some qualitative results of our method. We simply compare the descriptions generated by HOCA-U and L-HOCA-UBT, respectively. GT represents ``Ground Truth". Benefiting from the high-order correlation of multiple modalities, L-HOCA-UBT can generate more accurate descriptions which are close to GT.

%The first row is the results of video captioning and question answering. The ground-truth caption of the left video is "the boy is playing the cat". HOCA-UBT generates a description "a person is playing with a cat" which is similar to the ground truth. In terms of video question answering, HOCA-UBT can answer the question of the right video correctly, but the result of HOCA-U is wrong. The second and the third rows show the result of video summarization, we take the video "Smage Bros. Motorcycle Stunt Show" in TVSum as an example. The light-gray bars in the second row correspond to the ground-truth importance scores; while the red bars correspond to the selected parts of HOCA-UBT. The summary of HOCA-UBT is in accordance with the ground truth. The third row is several key frames of our summary, where the redundant frames are not included. 

%and the corresponding storage of the calculation process is also shown in Table \ref{table:size}.
%\vspace{-0.1cm}
%\vspace{0.5cm}

\section{Conclusion}

In this paper, we have proposed a new cross-modal attention mechanism called HOCA for video captioning. HOCA integrates the information of the other modalities into the inference of attention weights of current modality. Furthermore, we have introduced the Low-Rank HOCA which has a good scalability to the increasing number of modalities. The experimental results on two standard datasets have demonstrated the effectiveness of our approach.

\section*{Acknowledgments}
This work was supported in part by National Natural Science Foundation of China (No. 61702448, 61672456), Zhejiang Lab (2018EC0ZX01-2), the Fundamental Research Funds for the Central Universities (2019FZA5005), Artificial Intelligence Research Foundation of Baidu Inc, and Zhejiang University — HIKVision Joint lab.

%\subsection{Learning Curves}
%we show the learning curves of the CIDEr on the validation set in Fig. \ref{fig:curve}. We can observe that the L-HOCA-UBT performs better than HOCA-UBT and HOCA-U when the training converges.

%\begin{figure}[h]
%	\centering
%	\includegraphics[scale=0.45]{figs/training_curve.png}
%	\caption{Learning curves of different methods on MSR-VTT, where the rank of L-HOCA-UBT is 1. Note that we use greedy search during training while beam search during testing, so the testing scores are higher.}
%	\label{fig:curve}
%\end{figure}

\bibliography{emnlp-ijcnlp-2019}
\bibliographystyle{acl_natbib}

\clearpage

\appendix

\section{Supplemental Material}
\subsection{Proposition 1 for Eqn. \ref{13} in the paper}
{\bf Proposition 1: }Suppose that we have $n$ matrices, $I_1,I_2,...,I_n$, and $n$ vectors, $w_1, w_2,...,w_n$. The space of $I_l$ is $\mathbb{R}^{d \times t_l}$ and the space of $w_l$ is $\mathbb{R}^{1 \times t_l}$. Then

\vspace{-10pt}

\begin{equation}
\label{25}
\begin{aligned}
\Big (\bigotimes_{i=1}^n I_i \Big ) \!\circ \! \Big (\bigotimes_{i=1}^n w_i \Big ) \!= \! \bigotimes_{i=1}^n I_i \circ w_i 
\end{aligned}
\end{equation}

{\bf Proof: }We use $C_l$ and $C_r$ to denote the left side and right side of the equation, respectively. We utilize the element-wise comparison in two tensors. Following Eqn. \ref{7} and \ref{8} in the paper, the $(r_1,...,r_n)$-th entry of $C_{l}$ is expressed as

\vspace{-0.5cm}

\begin{equation}
\label{26}
(C_l)_{r_1,...,r_n} \!= \!{\bf 1}_d \Big [\myhardamard_{i=1}^n (I_i)_{r_i} \Big ] \cdot \Big [\myhardamard_{i=1}^n (w_i)_{r_i} \Big ]  
\end{equation}

\noindent where $(I_i)_{r_i}$ is a vector which denotes $r_i$-th column of the $I_i$, $(w_i)_{r_i}$ is the $r_i$-th value of the vector. Since $(w_i)_{r_i}$ is a single element, we can directly multiply it with the corresponding vector $(I_i)_{r_i}$.

\vspace{-10pt}

\begin{equation}
\begin{aligned}
\label{27}
(C_l)_{r_1,...,r_n} & \!= \!{\bf 1}_d \Big [\myhardamard_{i=1}^n (I_i \circ w_i)_{r_i} \Big ] \\
%(((I_1)_{r_1} \! \cdot \! (w_1)_{r_1}) \!\circ \!... \!\circ \! ((I_n)_{r_n} \!\cdot \!(w_n)_{r_n}) )\\
& = (C_r)_{r_1,...,r_n}
\end{aligned}
\end{equation}

The proposition is proven and is used to convert Eqn. \ref{12} to Eqn. \ref{13} in the paper.

\subsection{Proposition 2 for Eqn. \ref{15} in the paper}

{\bf Proposition 2: }Suppose that we have $n$ matrices, $I_1,I_2,...,I_n$. The space of $I_l$ is $\mathbb{R}^{d \times t_l}$. Then

\begin{equation}
\label{28}
\begin{aligned}
\sum \Big (\bigotimes_{i=1}^n I_i \Big ) =  {\bf 1}_d \Big [\myhardamard_{i=1}^n (I_i) {\bf 1}_{t_i} \Big ]
%((I_1){\bf 1}_{t_1} \circ ... \circ (I_n){\bf 1}_{t_n})
\end{aligned}
\end{equation}

\noindent here we use vectors ${\bf 1}_d$ and ${\bf 1}_{t_i}$ which consist of $1$ to represent the summation operation for matrix $I_i$ in $d$ dimension and $t_i$ dimensions, respectively.

{\bf Proof: }We use $v_l$ to denote the left side of the equation and $v_r$ to denote the right side of the equation. We can express $v_l$ as

\begin{equation}
\label{29}
\begin{aligned}
v_l =  \sum_{r_1 = 1}^{t_1} ... \sum_{r_n = 1}^{t_n} {C_{r_1,r_2,...,r_n}} 
\end{aligned}
\end{equation}

\begin{equation}
\label{30}
\begin{aligned}
C = \bigotimes_{i=1}^n I_i  
\end{aligned}
\end{equation}

Following Eqn. \ref{7} and \ref{8} in the paper, we can express $C_{r_1,r_2,...,r_n}$ as

\begin{equation}
\label{31}
{C_{r_1,r_2,...,r_n}} = {\bf 1}_d \Big [\myhardamard_{i=1}^n (I_i)_{r_i} \Big ]
%((I_1)_{r_1} \circ ... \circ (I_n)_{r_n})
\end{equation}

We apply Eqn. \ref{32} to Eqn. \ref{30},

\begin{equation}
\begin{aligned}
\label{32}
v_l &=  \sum_{r_1 = 1}^{t_1} ... \sum_{r_n = 1}^{t_n} {{\bf 1}_d \Big [\myhardamard_{i=1}^n (I_i)_{r_i} \Big ]} \\
&= {\bf 1}_d \Big[\sum_{r_1 = 1}^{t_1} ... \sum_{r_n = 1}^{t_n} { \myhardamard_{i=1}^n (I_i)_{r_i}} \Big ] \\
&= {\bf 1}_d \Big [\myhardamard_{i=1}^n (I_i) {\bf 1}_{t_i} \Big ] = v_r
\end{aligned} 
\end{equation}

%Following the equation given above, we can obtain the Eqn.\ref{166} in the paper.
The proposition is proven and is used to convert Eqn. \ref{13} to Eqn. \ref{15} in the paper.

\subsection{Learning Curves}
\begin{figure}[h]
	\vspace{-0.5cm}
	\centering
	\includegraphics[scale=0.5]{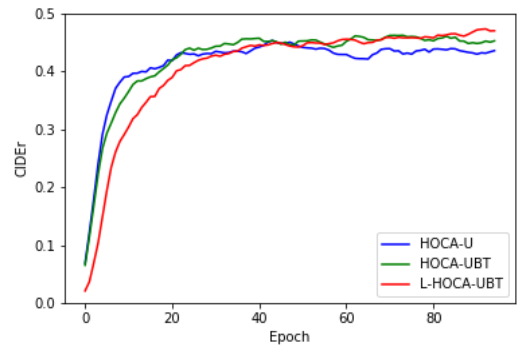}
	\caption{Learning curves of different methods on MSR-VTT, where the rank of L-HOCA-UBT is 1. Note that we use greedy search during training while beam search during testing, so the testing scores are higher.}
	\label{fig:curve}
	\vspace{-0.3cm}
\end{figure}

We show the learning curves of the CIDEr on the validation set in Fig. \ref{fig:curve} and observe that the L-HOCA-UBT performs better than HOCA-UBT and HOCA-U when the training converges.

\subsection{Visualization of Attention Weights}
\begin{figure}[t]
	\centering
	\includegraphics[scale=0.8]{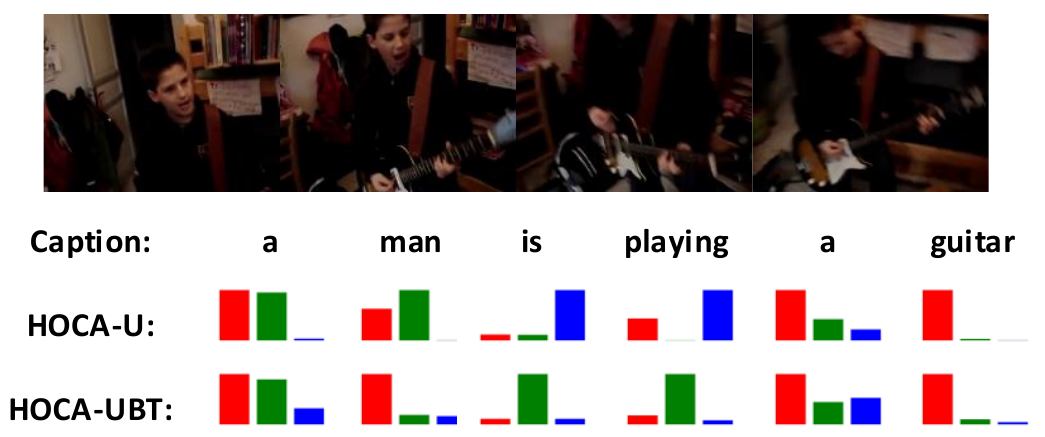}
	\caption{Visualization of the attention weights in multiple attentive fusion (MAF) module, the red bar denotes image modality, the green bar denotes motion modality, the blue bar denotes audio modality. }
	\label{fig:visual}
\end{figure}

We also perform visualization of the attention weights in multiple attentive fusion (MAF) module. As shown in Fig. \ref{fig:visual}, HOCA-UBT obtains a more accurate ratio of each modality than HOCA-U, i.e. for the word ``man", HOCA-U obtains a higher score of motion modality, which violates human subjective understanding.

\end{document}